\PassOptionsToPackage{dvipsnames,table}{xcolor}
\documentclass[sigconf=true, nonacm=true, review=false, anonymous = false]{acmart}
\usepackage{xcolor}
\usepackage{graphicx}
\usepackage{caption}
\usepackage{subcaption}
\usepackage{comment}
\AtBeginDocument{%
  \providecommand\BibTeX{{%
    \normalfont B\kern-0.5em{\scshape i\kern-0.25em b}\kern-0.8em\TeX}}}

\begin{document}

\title{Algorithm Instance Footprint: Separating Easily Solvable and Challenging Problem Instances}

\author{Ana Nikolikj}
\orcid{0000-0002-6983-9627}
\affiliation{
  \institution{Jo\v{z}ef Stefan Institute \&}
  \institution{Jo\v{z}ef Stefan International Postgraduate School}
  \streetaddress{Jamova cesta 39}
  \city{Ljubljana}
   \country{Slovenia}
  \postcode{1000 }
}

\author{Sa\v{s}o D\v{z}eroski}
\orcid{0000-0003-2363-712X}
\affiliation{%
  \institution{Jo\v{z}ef Stefan Institute \&} 
  \institution{Jo\v{z}ef Stefan International Postgraduate School}
  \country{Slovenia}
}

\author{Mario Andrés Muñoz}
\orcid{0000-0002-7254-2808}
\affiliation{
  \institution{OPTIMA, The University of Melbourne}
  \country{Australia}
}

\author{Carola Doerr} 
\orcid{0000-0002-5983-7169}
\affiliation{%
  \institution{Sorbonne Université, CNRS, LIP}
  \streetaddress{}
  \country{France}}

\author{Peter Koro\v{s}ec}
\orcid{0000-0003-4492-4603}
\affiliation{
  \institution{Jo\v{z}ef Stefan Institute}
  \country{Slovenia}
}

\author{Tome Eftimov}
\orcid{0000-0001-7330-1902}
\affiliation{
  \institution{Jo\v{z}ef Stefan Institute}
  \country{Slovenia}
}

\begin{abstract}
In black-box optimization, it is essential to understand why an algorithm instance works on a set of problem instances while failing on others and provide explanations of its behavior. We propose a methodology for formulating an algorithm instance footprint that consists of a set of problem instances that are easy to be solved and a set of problem instances that are difficult to be solved, for an algorithm instance. This behavior of the algorithm instance is further linked to the landscape properties of the problem instances to provide explanations of which properties make some problem instances easy or challenging. The proposed methodology uses meta-representations that embed the landscape properties of the problem instances and the performance of the algorithm into the same vector space. These meta-representations are obtained by training a supervised machine learning regression model for algorithm performance prediction and applying model explainability techniques to assess the importance of the landscape features to the performance predictions. Next, deterministic clustering of the meta-representations demonstrates that using them captures algorithm performance across the space and detects regions of poor and good algorithm performance, together with an explanation of which landscape properties are leading to it. 
\end{abstract}




\maketitle

\section{Introduction}
 Many algorithms for solving continuous single-objective optimization (SOO) problems have been created and their effectiveness has mostly been evaluated through statistical analysis~\cite{stork2020new}. The commonly used approaches of computing average performance across a set of benchmark problem instances~\cite{derrac2011practical} or comparing distributions for a chosen performance metric~\cite{eftimov2017novel} have been criticized for a long time~\cite{hall2010generation}. However, there is still a lack of available methodologies and tools to address this issue which means that these practices continue to be used. As a result, the scientific results from such comparisons often do not generalize to new problem instances hindering the progress toward trustworthy optimization and making it difficult to understand the behavior of algorithms. This practice decreases confidence in the use of the algorithms for solving new optimization problem instances.

 The primary issue in the direction of understanding algorithm behavior is that we have a limited understanding of the behavior of these algorithms and thus they are treated as black-box systems. The performance of an algorithm instance can vary significantly based on the optimization problem instances it is trying to solve. A deeper understanding of the interaction between the algorithm, the optimization problem, and the performance would allow us to identify properties that make a problem instance easy or challenging for a specific algorithm instance.

 \textbf{Our contribution:} We propose a methodology to understand how problem properties and algorithm performance interact. We use meta-representations, which integrate problem properties and algorithm performance into a single vector space, created by training a machine learning regression model and analyzing feature importance. Clustering the meta-representations using a deterministic approach reveals regions of good and poor algorithm performance that define the algorithm instance footprint. Post-hoc analysis of the regions identifies the problem properties causing it. This sheds light on the algorithm's behavior and provides insights into its strengths and weaknesses. Note that this methodology is not for comparing different algorithms but for understanding each algorithm's behavior.

\textbf{Related Work.} 
The most commonly-used practice in analyzing algorithm behavior is performing a performance assessment which relies on statistical analysis, to compare performance data of different algorithm instances across a selected set of benchmark problem instances~\cite{derrac2011practical,eftimov2017novel}. The main drawback in such comparisons is that sometimes algorithm instances can be the best for some problem instances but worse for another set of problem instances, which actually affects the final statistical results. Such comparison results do not provide explanations for which algorithm instance is suitable for which problem instance and why.

Instance Space Analysis (ISA)~\cite{munoz2017performance,smith-miles2022instance} has been introduced to understand the intricate relationships between the algorithm instance behavior and the problem instance properties. It identifies regions in the problem space where a specific algorithm performs well and solves problems easily, by linking problem landscape features with a single performance metric. For this purpose, ISA uses a dimension reduction technique that encourages linear trends to appear both in the features and algorithm performance distributions to support visualizations. Then, it uses a supervised classifier to identify the areas in which the evidence of good performance is the strongest, defined as whether the instances are solvable or not within some precision of the global optima. The areas are characterized by their size, density, and purity, giving metrics of the algorithm's strength relative to the diversity of the benchmarks. Therefore, ISA represents an improvement in comprehending algorithm behavior, describing complex interactions in a linear manner.

Another way to understand the interactions between algorithm instances and problem instances is through the application of supervised regression models for algorithm configuration (finding the best hyper-parameters of an algorithm instance)~\cite{belkhir2017per,de2022improved} and algorithm selection (selecting the best algorithm for a given problem instance)~\cite{kerschke2019automated,jankovic2021towards,kostovska2022per}. In this approach, algorithm performance is predicted based on the features extracted from the problem instance landscape, which serve as problem-instance meta-representations. Most studies in this field use meta-learning~\cite{vanschoren2019meta} to train a single regression model to predict the performance of an algorithm instance on all problem instances. Recent studies ~\cite{trajanov2021explainable,trajanov2022explainable} have explored an explainable workflow for automated algorithm performance prediction. They utilized feature importance analysis to identify the crucial landscape features that predict an algorithm's instance performance on a global scale (i.e., a set of benchmark problem instances) and a local scale (i.e., a specific problem instance).

\vspace{-2mm}

\section{Algorithm Instance Footprint}
\label{sec:footprint}
Let us assume that we have a set of benchmark problem instances represented by their landscape features and linked to the performance of an algorithm instance achieved on them. The set is further split into train and test data sets. The term "algorithm instance footprint" refers to the regions (i.e., sets of problem instances) where an algorithm instance performs well or poorly, with accompanying identification of the problem landscape properties that contribute to this performance variation. The methodology for generating it consists of four steps (see Figure~\ref{fig:flowchart}):

\begin{figure*}
   \centering
   \includegraphics[scale=0.6]{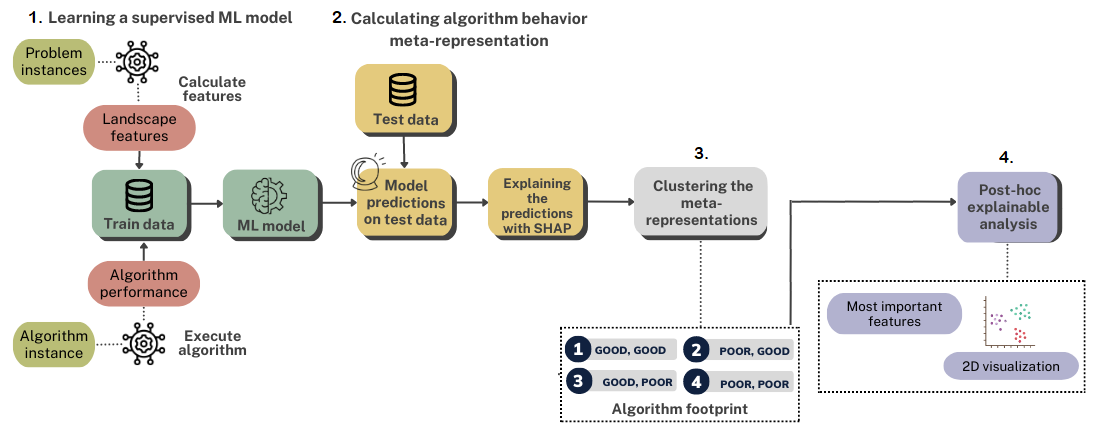}
   \caption{Flowchart of the methodology for calculating and analyzing algorithm instance footprint.}
   \label{fig:flowchart}
\end{figure*}

\noindent\textbf{1) Learning a supervised ML model} -- Train a supervised ML regression model with the train data, which will predict the performance of the algorithm instance for each problem instance based on its landscape feature representation. Use this model to predict the algorithm instance's performance on the test data set.

\noindent\textbf{2) Calculating algorithm behavior meta-representations} -- Once the predictions have been made on the test data set, we apply explainable techniques to assess the importance of the landscape features that impact the algorithm's performance prediction for each problem instance independently. These contributions can be used as a meta-representation of the algorithm's behavior on a specific problem instance. We have opted for the SHAP method~\cite{rozemberczki2022shapley} since it offers localized explanations that incorporate both the problem instance's landscape features and the algorithm instance's performance.

\noindent\textbf{3) Clustering the meta-representations} --  
By computing meta-representations that depict the algorithm instance's behavior on each problem instance, we can categorize the problem instances into two sets: those that are easy for the algorithm instance to solve, and those that are challenging. We have established a deterministic approach to accomplish this, resulting in more straightforward findings. The meta-representations are clustered into four groups using deterministic clustering, based on two factors: i) poor or excellent optimization algorithm instance performance and ii) poor or excellent ML prediction. In both situations, apriori set thresholds are employed to differentiate between poor and excellent performance. For ground truth performance, this implies that an error of no more than a predetermined target $t$ is required for excellent algorithm performance. For the ML model, it is determined that excellent predictive performance is achieved when the error is within $p\%$ of the predicted true precision. Below are explanations of the clusters.

           \textbf{(Good, Good):} This pertains to the depiction of algorithmic behavior, which identifies the problem instances where the algorithm instance's discovered solution quality is high, and the ML model accurately predicts the performance with a negligible error. This situation arises when the optimization algorithm easily solves a problem instance, and the ML algorithm recognizes this behavior.

            \textbf{(Poor, Good):} This involves behavior observed on problem instances where the algorithm instance's solution quality is low, yet the ML algorithm correctly predicts it with minimal error. This scenario arises when a problem instance proves to be challenging for the optimization algorithm instance to solve, and the ML algorithm identifies this behavior.
        
            \textbf{(Good, Poor):} This identifies the problem instances where the algorithm instance's discovered solution quality is high, but the ML algorithm failed to predict it.
        
            \textbf{(Poor, Poor):} This encompasses problem instances  where the algorithm instance's solution quality is low and the ML algorithm failed to predict this behavior.

\noindent\textbf{4) Post-hoc explainable analysis} -- Once the clusters have been identified, a post hoc explainable analysis is performed. This analysis clarifies which landscape features make the problem instances easily solved or challenging to solve for the algorithm instance. It is delivered by identifying the most important features of each of the clusters from the previous step, and also by providing 2D visualization of the meta-representations in which the algorithm performance and the most important feature values are visualized across the space.
\vspace{-2mm}

\section{Experimental design}
\label{sec:experimental_design}
\label{sec:experimentalsetup}

\noindent\textbf{Problem portfolio.}
The study uses the BBOB (i.e., COCO) benchmark suite~\cite{hansen2010real,hansen2021coco}. It features 24 noise-free, single-objective optimization problems, which can be altered by scaling and translating in the objective space to create different instances. This work uses the first five instances of each problem, totaling 120 benchmark instances. The problem dimension is set to 10, $D=10$.

\noindent\textbf{Landscape features.}
We select the most commonly used landscape features that are used to describe the properties of single-objective optimization problems, known as ELA features~\cite{mersmann2011exploratory}. Their calculation has been taken from a previous study~\cite{lang2021exploratory}. 
A total of 64 features were selected, including classical ELA features~\cite{mersmann2011exploratory}, Dispersion~\cite{lunacek2006dispersion}, Information Content~\cite{munoz2014exploratory}, Nearest Better Clustering~\cite{kerschke2015detecting}, and Principal Component Analysis~\cite{kerschke2019comprehensive}.

\noindent\textbf{Algorithm portfolio.}
Three randomly selected Differential Evolution (DE) configurations have been selected as the algorithm portfolio just to present how the proposed methodology works. The configurations (DE1, DE2, and DE3) are taken from a previous study~\cite{nikolikj2022explaining}, where each configuration is presented in more detail including its strategy, $F$, and $Cr$ values.

\noindent\textbf{Performance data.} 
The study focuses on the fixed-budget performance scenario, where the target precision of the algorithm (i.e., the distance between the best solution and the estimated optimal solution) is used as a performance metric. The logarithm of the precision is calculated to capture the distance level to the optimum~\cite{jankovic2021towards}. The budget has been set on 500$D$ function evaluations. Each configuration has been run 30 times and the median reached precision is used as an approximation of its performance.

\begin{figure}[htb]
\centering
\begin{subfigure}[b]{0.35\textwidth}
   \includegraphics[width=0.9\linewidth]{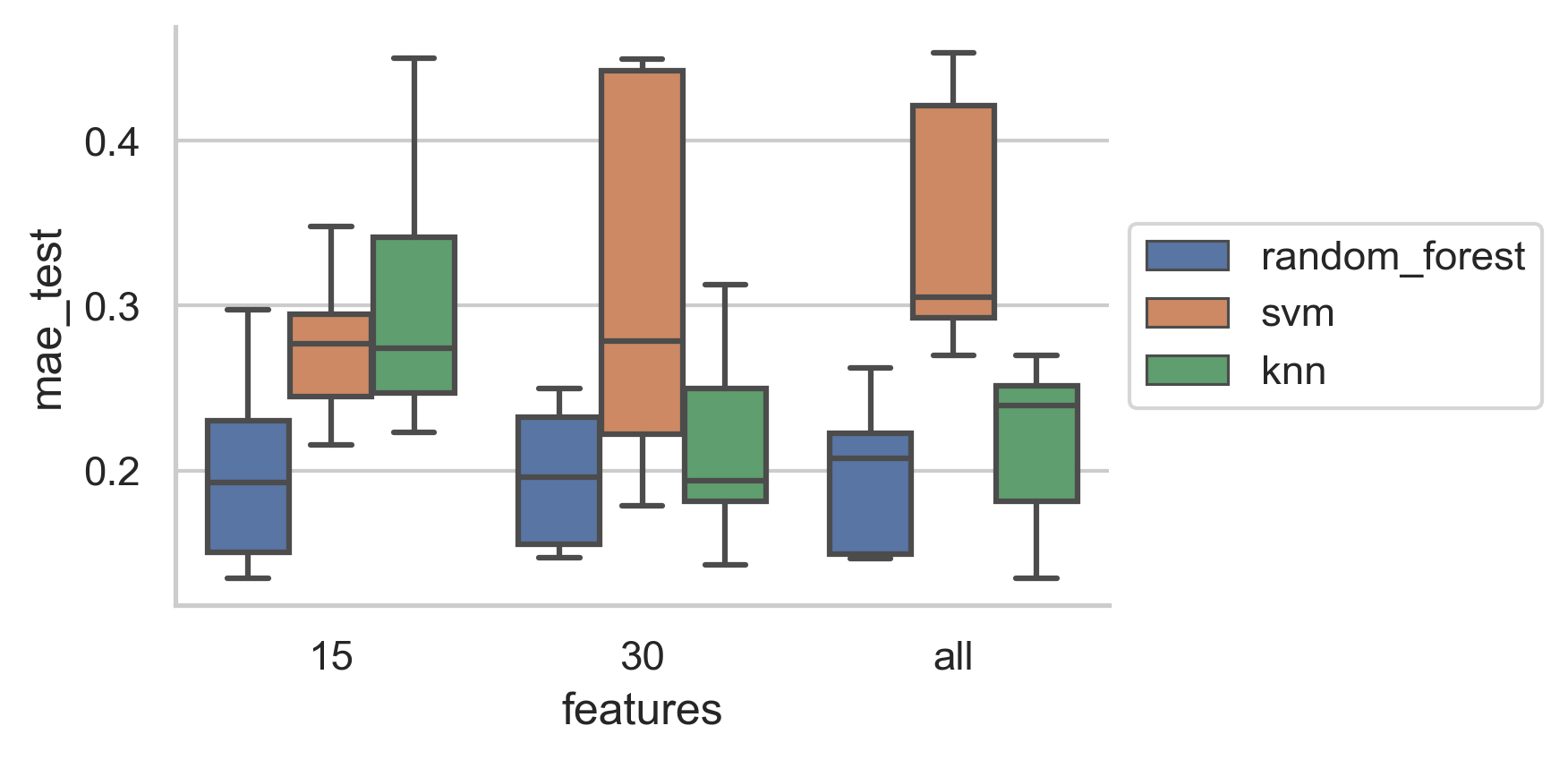}
   \vspace{-2mm}
   \caption{MAE}
   \label{fig:mae} 
\end{subfigure}

\begin{subfigure}[b]{0.35\textwidth}
   \includegraphics[width=0.9\linewidth]{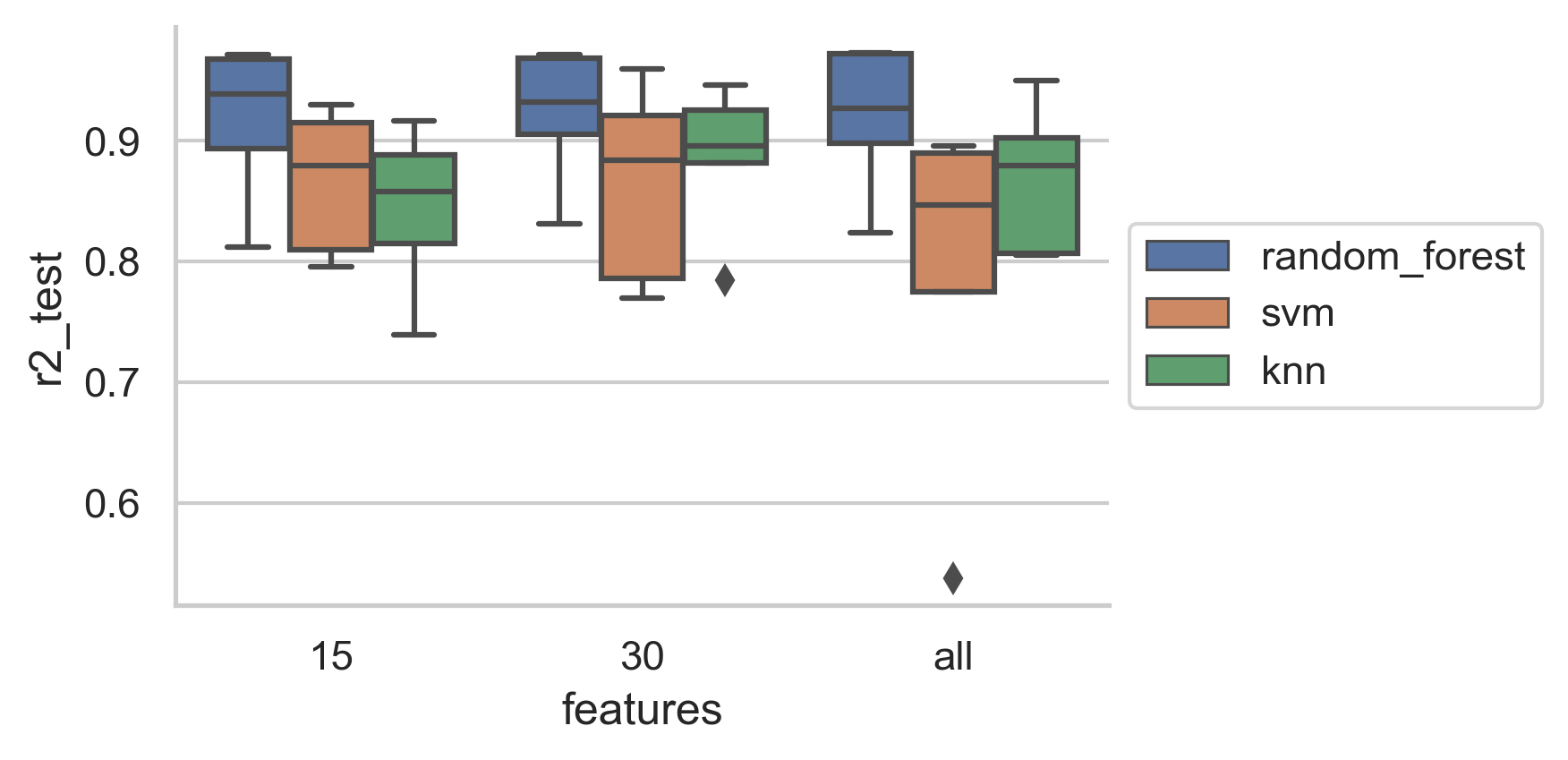}
   \vspace{-2mm}
   \caption{R2score}
   \label{fig:r2score}
\end{subfigure}
\vspace{-3mm}
\caption{Box-plot showing the distribution of model performance over the test portion of the five folds: (a) MAE, (b) R2 score, for different feature portfolios of most important features as identified by the SHAP method, when predicting the performance of DE$1$. }
\label{fig:performance}
\end{figure}

\noindent\textbf{Predictive models.} To find a good-performing supervised regression model, we evaluate three different regression families: Random Forest~\cite{biau2016random}, Support Vector Machines (SVM)~\cite{noble2006support}, and K-Nearest Neighbours (KNN)~\cite{peterson2009k}. Each one has been tested with different feature portfolios selected by the SHAP method~\cite{rozemberczki2022shapley}. To select the feature portfolio, first, a model is trained with all features, then the SHAP feature importance is calculated, and finally the top most important features are selected as indicated by the Shapely scores. The models have been evaluated in stratified five-fold cross-validation, where one instance from each problem has been left for testing and the others four remain in the training data. Box plots depicting model performance on the test data (as measured by MAE and R2 score) for all regression models and different feature portfolios across the five folds are shown in Figure~\ref{fig:performance}. 
 The box plots demonstrate that the RF is robust to the different folds, while the performance of KNN and SVM appears more variable. The result is consistent across all feature portfolios. Using the results, we decide to use the RF model with 30 ELA features to show how the methodology for generating the algorithm footprint works. We need to highlight here that the footprint is calculated five times, for each fold separately. This allows us to investigate the robustness of the algorithm instance performance on the transformations (e.g., shifting or scaling) applied to generate different problem instances.
 
\vspace{-5mm}
\section{Results and discussion}
\label{sec:results}
\begin{table*}[!htb]
\caption{Distribution of the BBOB problem instances across the deterministic clusters for each fold.}
\label{t:model_performance}
\centering
\vspace{-2mm}
\resizebox{.85\textwidth}{!}{%
\begin{tabular}{ccllll}
\hline
model & fold number &     (good, good) &  (good, poor) & (poor, good) &   (poor, poor) \\
\hline
RF & 1 & 16, 19, 20, 21, 22 & 1, 2, 5, 14, 17, 18, 23 & 3, 4, 6, 7, 8, 9, 10, 11, 12, 15, 24 & 13 \\
RF & 2 & 19, 20, 21 & 1, 2, 5, 14, 17, 22, 23 & 3, 4, 6, 7, 8, 9, 10, 11, 12, 13, 15, 16, 24 & 18 \\
RF & 3 & 19, 20, 21, 22 & 1, 2, 5, 14, 16, 17, 18, 23 & 3, 4, 6, 8, 9, 12, 13, 15, 24 & 7, 10, 11 \\
RF & 4 & 5, 16, 18, 19, 20, 21, 22 & 1, 2, 7, 14, 17, 23 & 3, 4, 6, 8, 9, 10, 12, 13, 15, 24 & 11 \\
RF & 5 & 19, 20, 21 & 1, 2, 5, 7, 14, 16, 17, 22, 23 & 6, 8, 9, 11, 12, 13, 15, 24 & 3, 4, 10, 18 \\
\hline
\end{tabular}%
}
\end{table*}

To present how the proposed methodology can be used for understanding algorithm instance behavior, we select one DE configuration (DE1) for which we show the results and their interpretation in detail, while the results for the other two configurations (DE2 and DE3) are available at our GitHub repository~\cite{gitFootprints}, due to the page limits. We fix a target precision, $t$, to the median precision calculated over the training problem instances, to define if the algorithm instance can solve or not the problem instance within it, and a percentage of $p=15$\% that defines if an ML predictive model provides a good prediction within 15\% error. We need to point out here that these values ($t$ and $p$) are chosen as such only for illustration purposes and should be appropriately set according to the scenario in which the proposed methodology will be used. For example, the $t$ value should be set according to the acceptable optimization accuracy for the specific problem instance and the $p$ value should be set according to the acceptable model prediction accuracy. The acceptable model prediction accuracy depends on the application being solved, which are the tolerance levels for which a good prediction is acceptable. In this way, the methodology can be used to investigate the algorithm instance's strengths and weaknesses as required by the user.

\begin{figure*}
\begin{subfigure}{.39\textwidth}
  \centering
  % include first image
  \includegraphics[width=.95\linewidth]{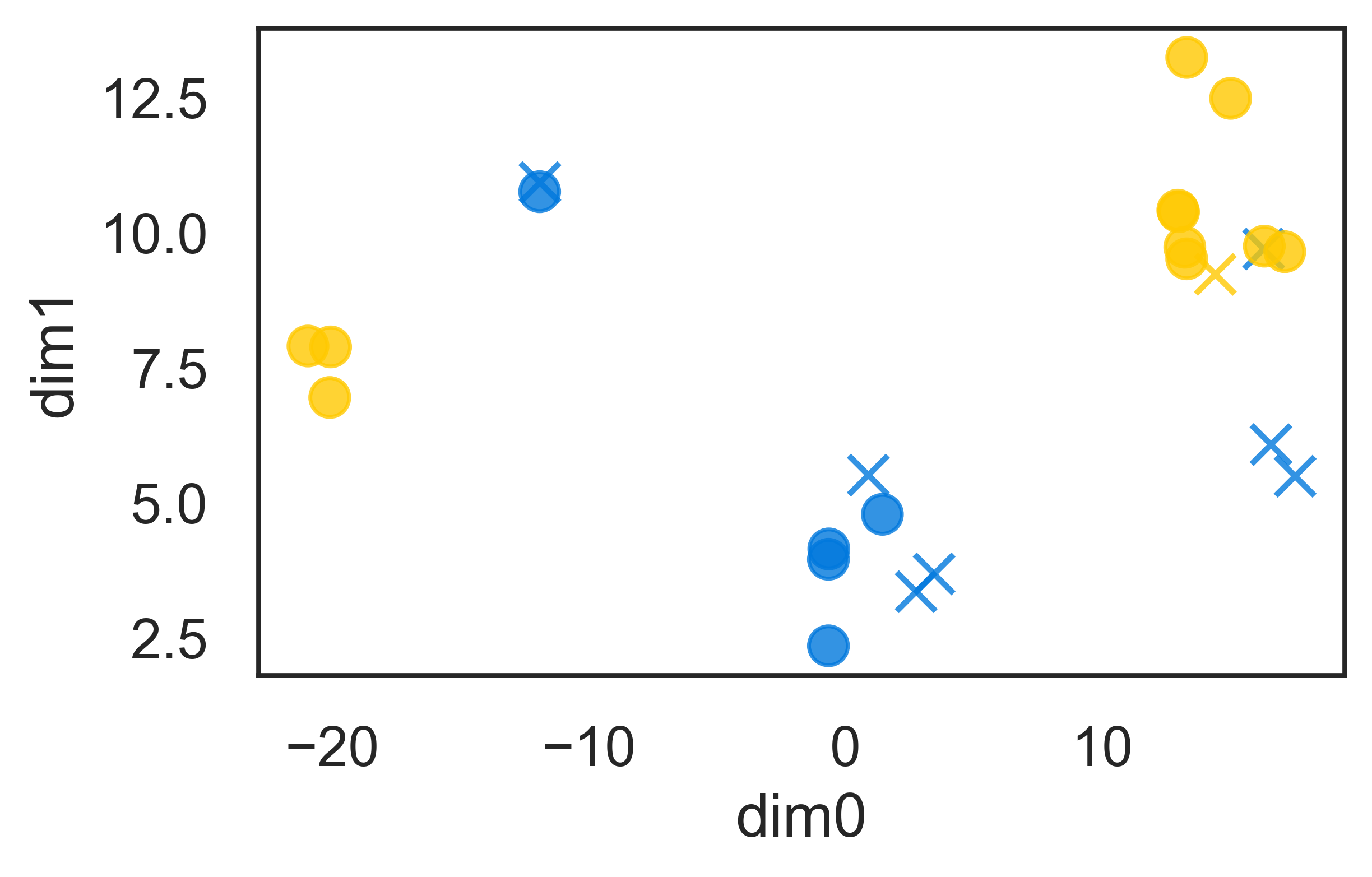}  
  \vspace{-2mm}
  \caption{First fold.}
  \label{fig:sub-first}
\end{subfigure}
\begin{subfigure}{.39\textwidth}
  \centering
  % include third image
  \includegraphics[width=.95\linewidth]{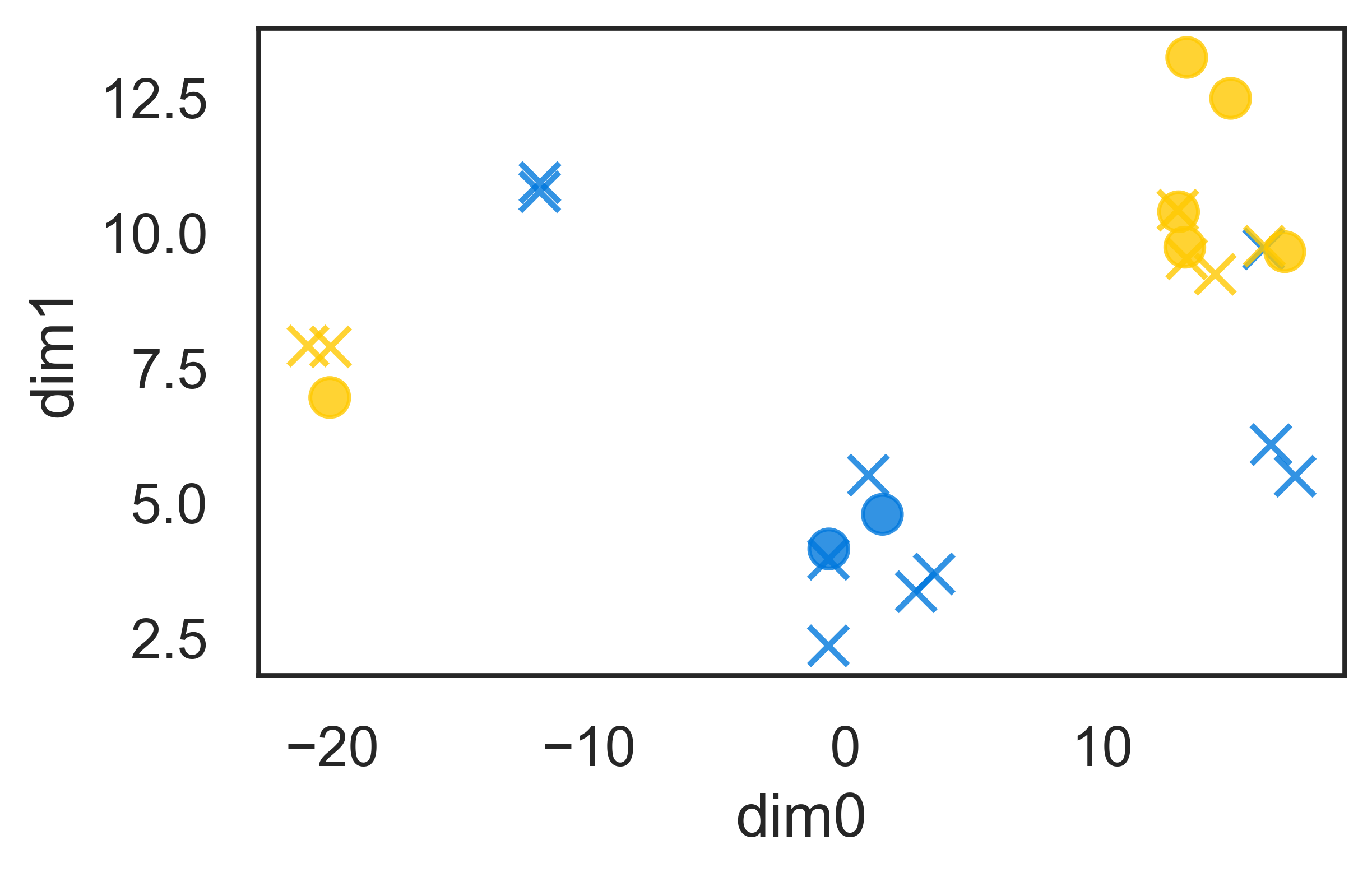}  
  \vspace{-2mm}
  \caption{First fold with threshold 5\% for RF error.}
  \label{fig:sub-sixt}
\end{subfigure}

\leavevmode\newline
\begin{subfigure}{.39\textwidth}
  \centering
  % include second image
  \includegraphics[width=.95\linewidth]{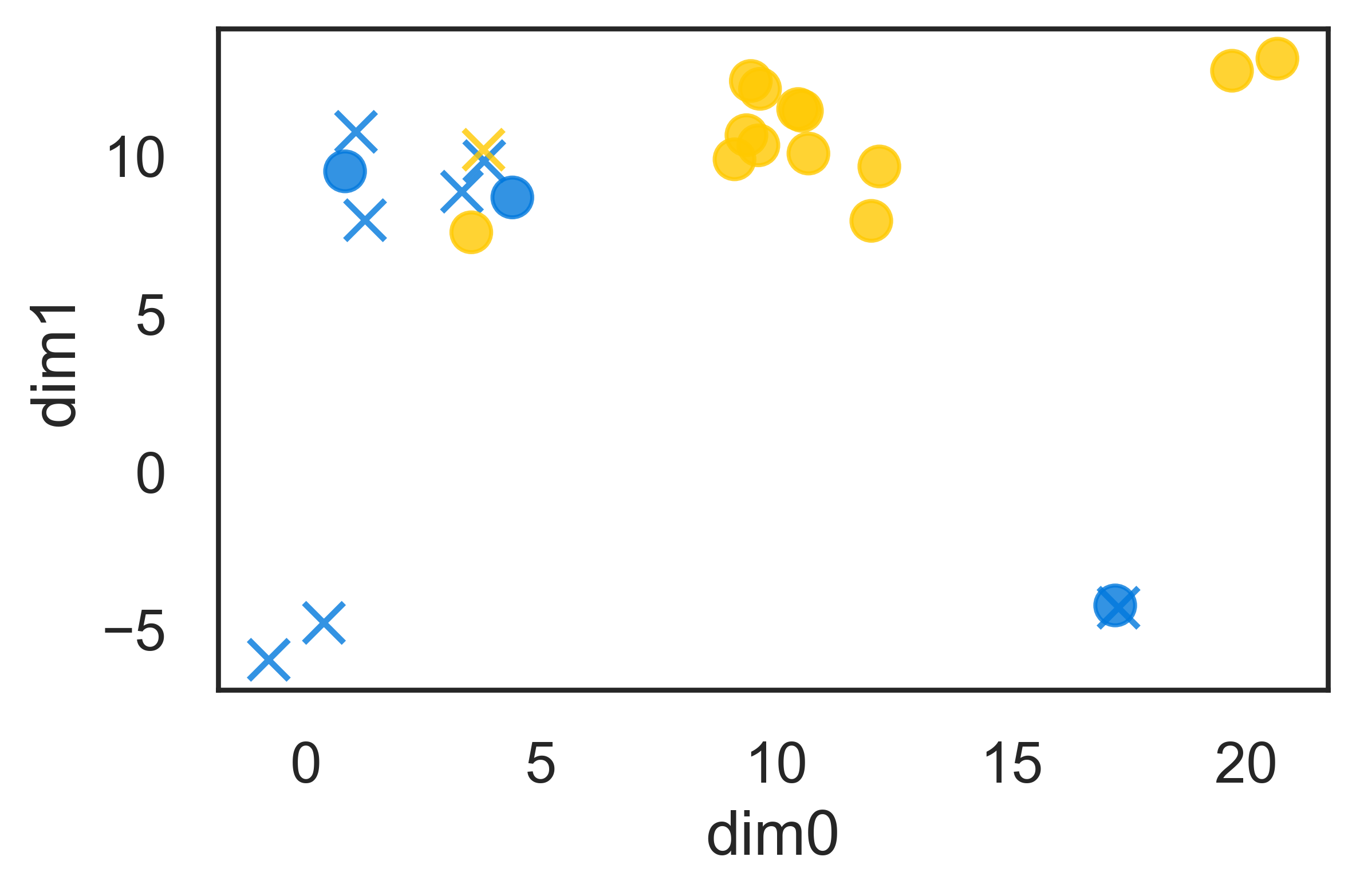}  \vspace{-2mm}
  \caption{Second fold.}
  \label{fig:sub-second}
\end{subfigure}
\begin{subfigure}{.39\textwidth}
  \centering
  % include third image
  \includegraphics[width=.95\linewidth]{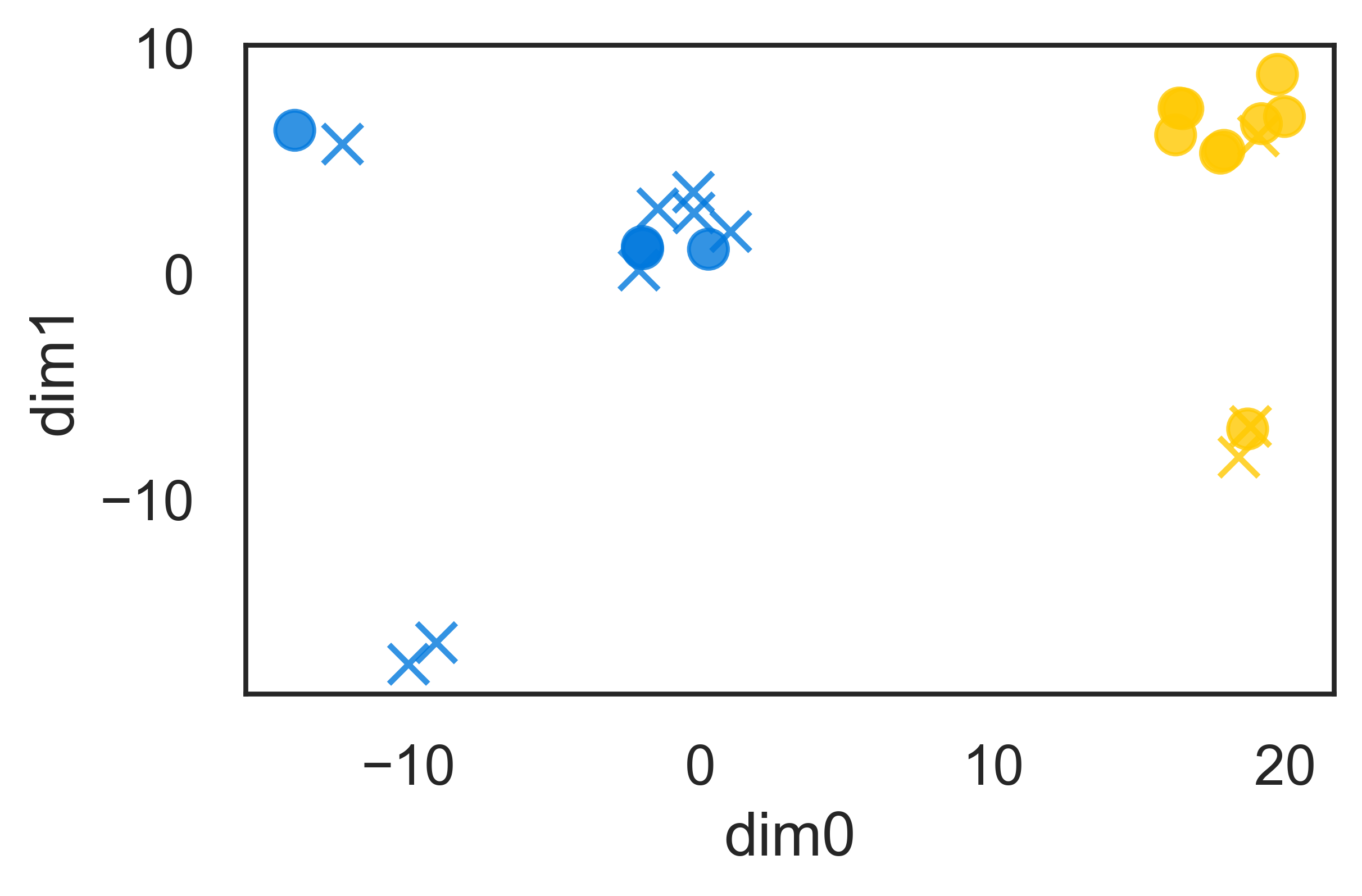}  \vspace{-2mm}
  \caption{Third fold.}
  \label{fig:sub-third}
\end{subfigure}

\leavevmode\newline
\begin{subfigure}{.39\textwidth}
  \centering
  % include the fourth image
  \includegraphics[width=.95\linewidth]{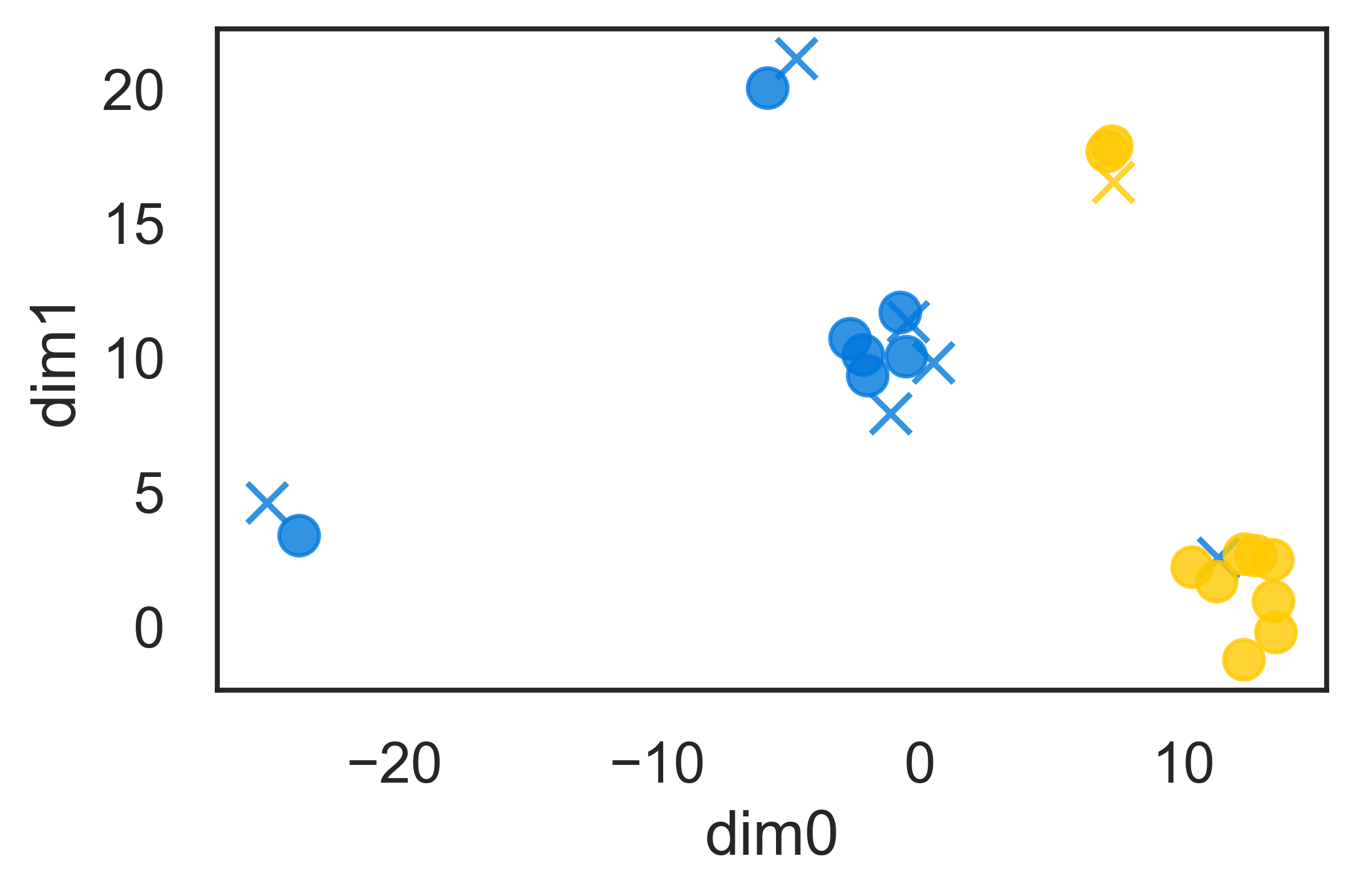}  \vspace{-2mm}
  \caption{Fourth fold.}
  \label{fig:sub-fourth}
\end{subfigure}
\begin{subfigure}{.39\textwidth}
  \centering
  % include the third image
  \includegraphics[width=.95\linewidth]{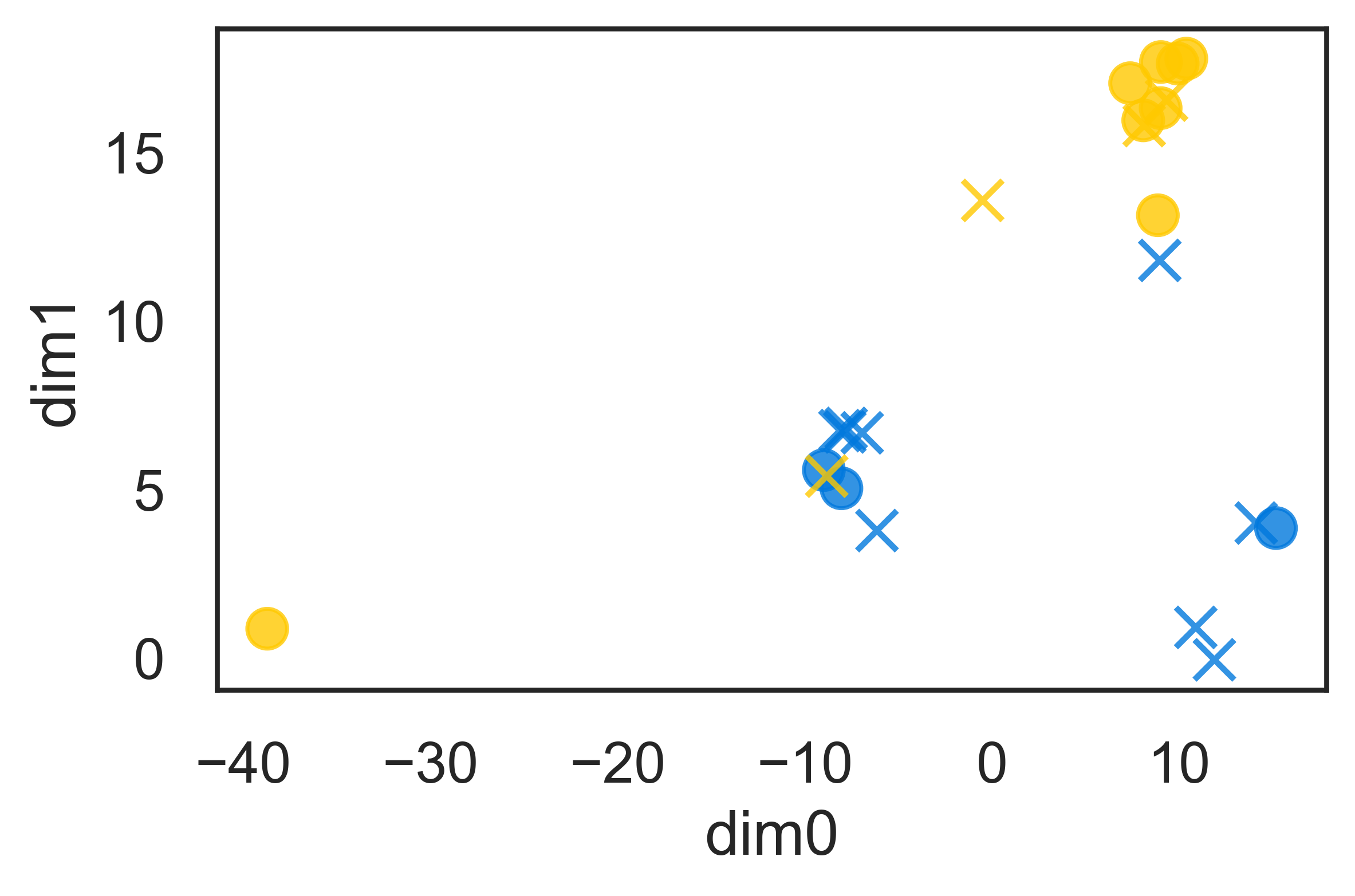}  \vspace{-2mm}
  \caption{Fifth fold.}
  \label{fig:sub-fifth}
\end{subfigure}

\leavevmode\newline
\vspace{-2mm}
\begin{subfigure}{.39\textwidth}
  \centering
  % include the fourth image
  \includegraphics[width=.65\linewidth]{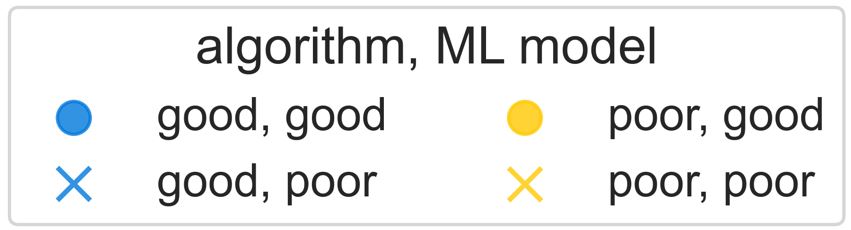}  
  \label{fig:legend}
\end{subfigure}

\vspace{-2mm}
\caption{2D UMAP ~\cite{mcinnes2018umap} visualization of the algorithm footprints obtained with the deterministic clustering, on the test portion of each of the five folds. The tolerance error for the RF model is within 15\%. The \textit{\textcolor{blue}{blue}} color represents regions of good algorithm performance, and the \textit{\textcolor{yellow}{yellow}} to regions of poor algorithm performance. 
The marker shape corresponds to good (O) and poor (X) ML model performance as indicated by the legend at the bottom of the plot.}
\label{fig:footprint}
\end{figure*}

\noindent\textbf{DE1 footprint.} Figure~\ref{fig:footprint} presents the 2D visualization of the DE1 footprint, for each fold separately. The footprints consist of four deterministic clusters that are obtained by pairing the ground truth algorithm instance performance and the ML performance. The most interesting are the following two combinations $(good, good)$ and $(poor, good)$. For the problem instances that belong to $(good, good)$ the algorithm instance solves them within the specified target $t$, and the ML predicts this behavior with an error of $p=15$\%. In the case of $(poor, good)$ the algorithm instance cannot solve the problem instances within the specified target, however, the ML model predicts its behavior within an error of 15\%. For these two combinations, we can provide additional explanations for the algorithm behavior. For the $(good, good)$, we can see which landscape features make those problem instances easily to be solved, while in the case of $(poor, good)$ which landscape features make those problem instances challenging to be solved. In the case of $(good, poor)$ and $(poor, poor)$, the ML model cannot predict if the algorithm is able to solve the problem instances within the specified target ($good$) or not ($poor$) within an error of 15\%. Because of this, for these two combinations, we are not able to find which landscape features are relevant to the algorithm behavior.

Table~\ref{t:model_performance} presents the distribution of the BBOB problem instances across the four clusters for each fold separately. From the results, we can conclude that DE1 has stable performance on the 19th (i.e., Composite Griewank-Rosenbrock Function), 20th (i.e., Schwefel Function), and 21st (i.e., Gallagher's Gaussian 101-me Peaks Function) BBOB problem classes. This comes from the fact that no matter the different transformations (e.g., shifting, scaling) that are applied to generate a problem instance of those problems that are parts of different folds, the algorithm instance is able the find a solution with the specified target and the RF model can predict it within an error of 15\%. For the 6th, 8th, 9th, 12th, 15th, and 24th BBOB problem classes (please find their names in~\cite{finck2010real}), the algorithm instance is not able to solve them within the specified target, however, the RF model can predict this behavior within an error of 15\%. This result indicates that no matter which transformations are applied to the base problem class to define the problem instances across different folds, the algorithm instance is not able to solve them. In the case of the $(good, poor)$ cluster, the algorithm instance can solve the problem instances from the 1st, 2nd, 14th, 17th, and 23rd problem classes, however, the RF model cannot predict this behavior across all folds.

There are also some interesting problem classes that depending on the transformations used to define the problem instances across different folds, the problem instances change their clusters. For example, the first, third, and fourth problem instances from the 22nd BBOB problem belong to the $(good, good)$ cluster. However, the second and the fifth instances of the same problem are presented in the $(good, poor)$ cluster. This result indicates that the algorithm instances for all problem instances can find a solution within the specified target. The difference is that the RF model cannot predict this behavior for the second and the fifth instance within an error of 15\% (the ML errors are 45\% and 17\% for the 2nd and 5th instances respectively.). Another example is the 5th BBOB problem class. Except for the fourth problem instance from this problem class, for all the remaining the RF error is greater than 15\% (for the fourth one is 8\%). 

Other examples contain the transition from $(poor, good)$ to $(poor,\\ poor)$: the 3rd, 10th, 11th, and 13th BBOB problem classes. All these problems correspond to problem instances that DE configuration is not able to solve within the specified target. The main difference is the performance of the RF model. For the 11th problem, the RF model can predict the behavior for the first, second, and fifth instances, while for the third and fourth is not able to predict it (i.e., RF errors above 15\%). This means that in the case of the first, second, and fifth instances we can further provide explanations of which landscape features make them difficult to be solved, while we cannot provide explanations for the third and fourth instances. Further, by performing an explainable post-hoc analysis, if different features are important between different folds for this problem class, this indicates that the important features for the third and fourth problem instances are ``wrong" features that are leading to poor prediction. If there is no difference between the important features across the folds, it means that the RF predictive model does not have enough power (i.e., confidence) to provide explanations for the third and fourth problem instances. Similar explanations are also present for the 3rd (ML error is 17\% for the fifth instance of this problem), 10th (RF errors are 18\% and 44\% for the third and fifth instances of this problem), and 13th (i.e., the RF error for the first instance of this problem is around 17\%) problem.

Some of the transitions from $(good, good)$ to $(good, poor)$ and vice-versa, and also from $(poor, good)$ to $(poor, poor)$ and vice versa, are happening only when the RF error is in some close $\epsilon$-neighborhood with the selected percentage, $p=15\%$. 

The problem instances on the 7th and the 18th problem classes are distributed across most of the clusters. This result points out that the algorithm instance does not have stable performance on them. It is either able to solve or not the problem instance within the specified target, which further can be a problem for the ML model to make a prediction. 

In general, analyzing the footprints across all folds, we can conclude that the footprints make a clear distinction between \textit{good} vs. \textit{poor} algorithm instance performance (i.e., placing $(good, good)$ to $(good, poor)$ problem instances together vs. $(poor, good)$ to $(poor,\\ poor)$ together). The second dimension, which is the ML model performance, only guarantees confidence in providing further explanations for problem instances that are predicted in the tolerance error.

\noindent\textbf{Sensitivity analysis with regard to the tolerance percentage of the ML model error.} To see the influence of the selected percentage for the RF model error, in Figure~\ref{fig:sub-sixt} we present the same footprint as in Figure~\ref{fig:sub-first} generated on the data from the first fold with a difference that the percentage for the RF model error is set at $p=5\%$. As was expected, the main transitions only happen from  $(good, good)$ to $(good, poor)$, and from $(poor, good)$ to $(poor, poor)$. If the target precision for the true algorithm instance increases and the RF model is fixed, the possible transitions are from $(good, good)$ to $(poor, good)$, and from $(good, poor)$ to $(poor, poor)$, otherwise, the transition is in the other direction.

\begin{figure*}[!htb]
\begin{subfigure}{.4\textwidth}
  \centering
  % include the first image
  \includegraphics[width=.95\linewidth]{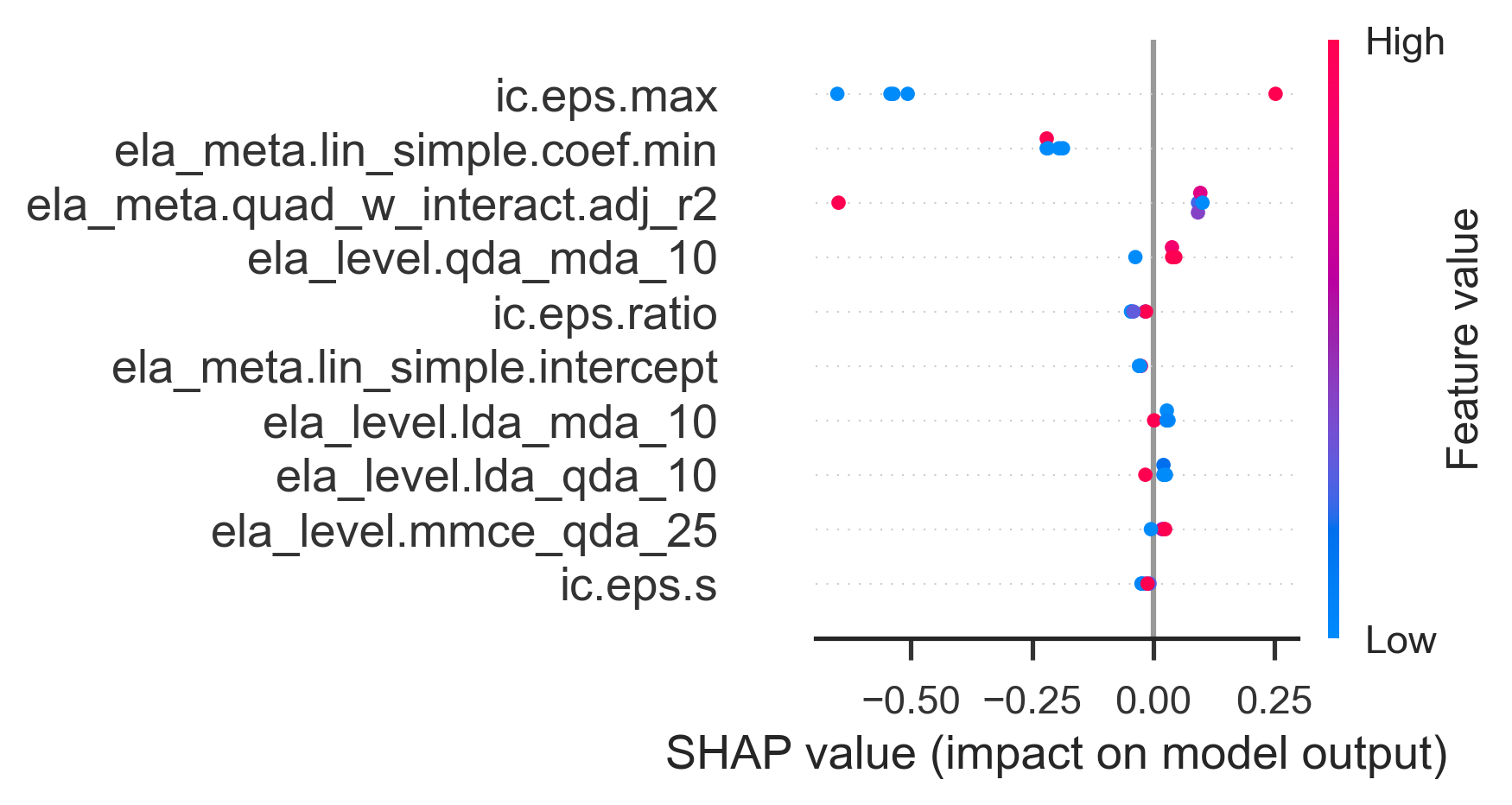}  
  \caption{First fold (good, good).}
  \label{fig:sub-first-shap}
\end{subfigure}
\begin{subfigure}{.4\textwidth}
  \centering
  % include the second image
  \includegraphics[width=.95\linewidth]{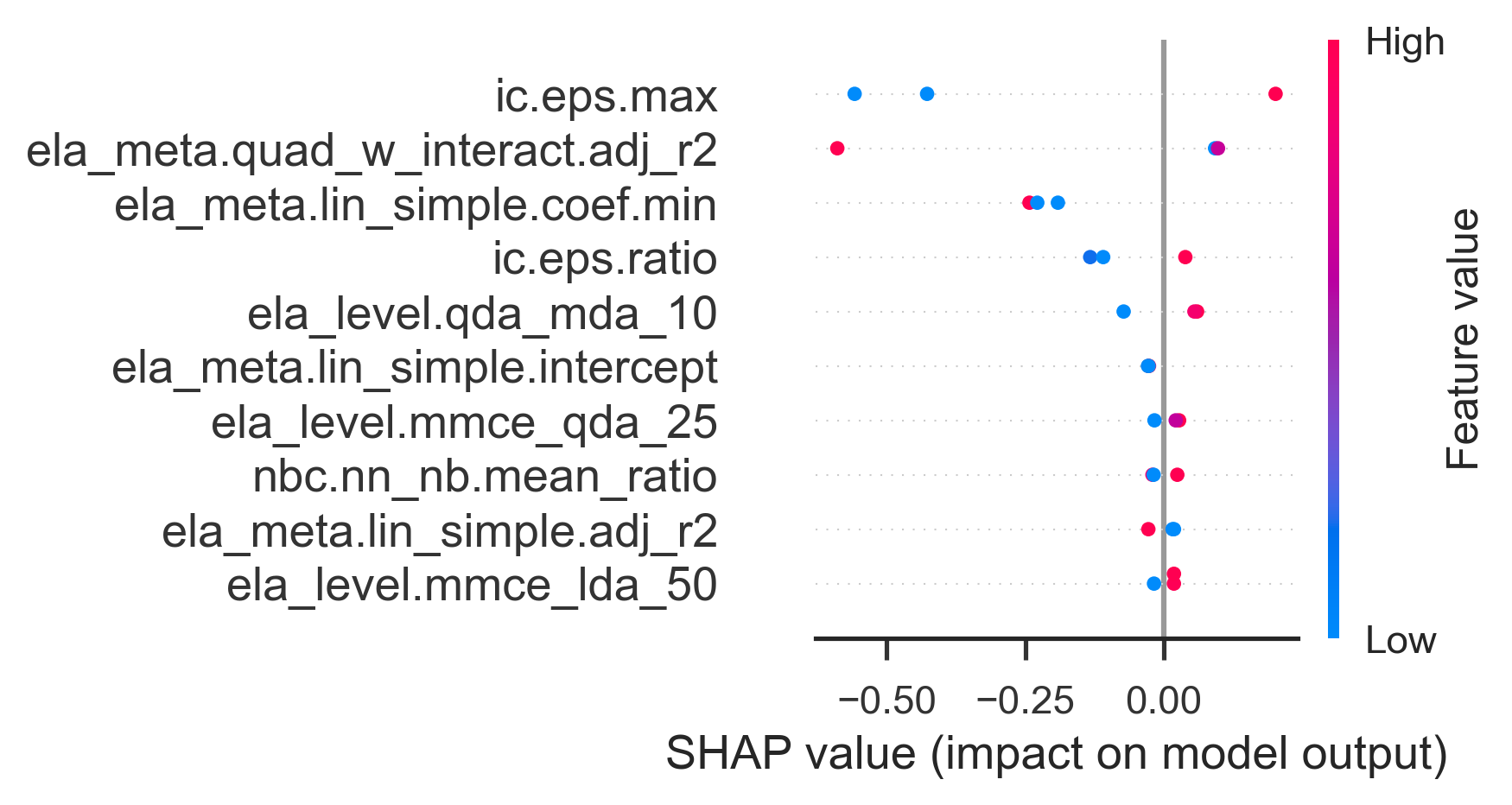}  
  \caption{Second fold (good, good).}
  \label{fig:sub-second-shap}
\end{subfigure}
\leavevmode \newline
\vspace{-2mm}
\begin{subfigure}{.4\textwidth}
  \centering
  % include the third image
  \includegraphics[width=.95\linewidth]{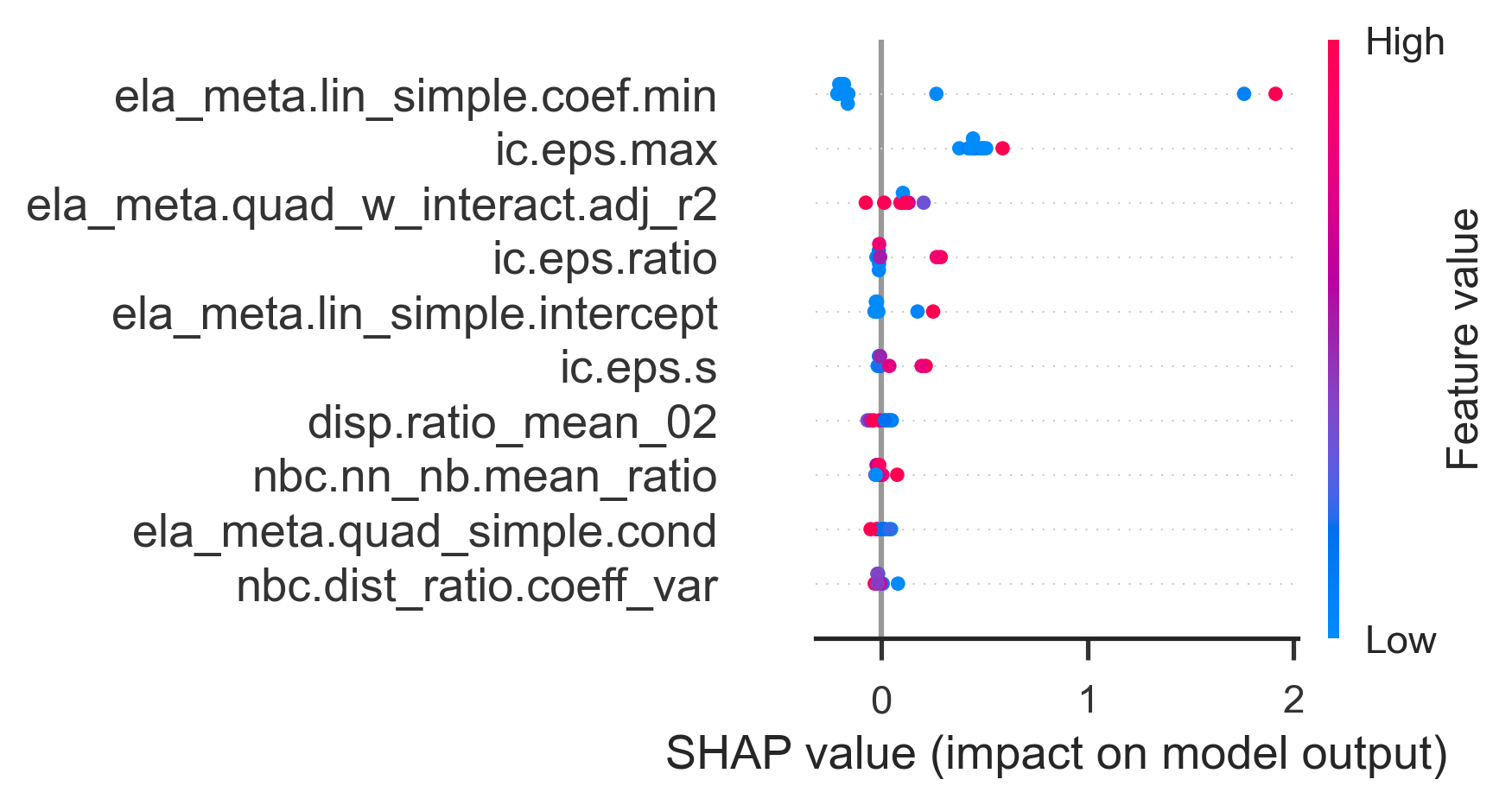}  
  \caption{First fold (poor, good).}
  \label{fig:sub-third-shap}
\end{subfigure}
\begin{subfigure}{.4\textwidth}
  \centering
  % include the fourth image
  \includegraphics[width=.95\linewidth]{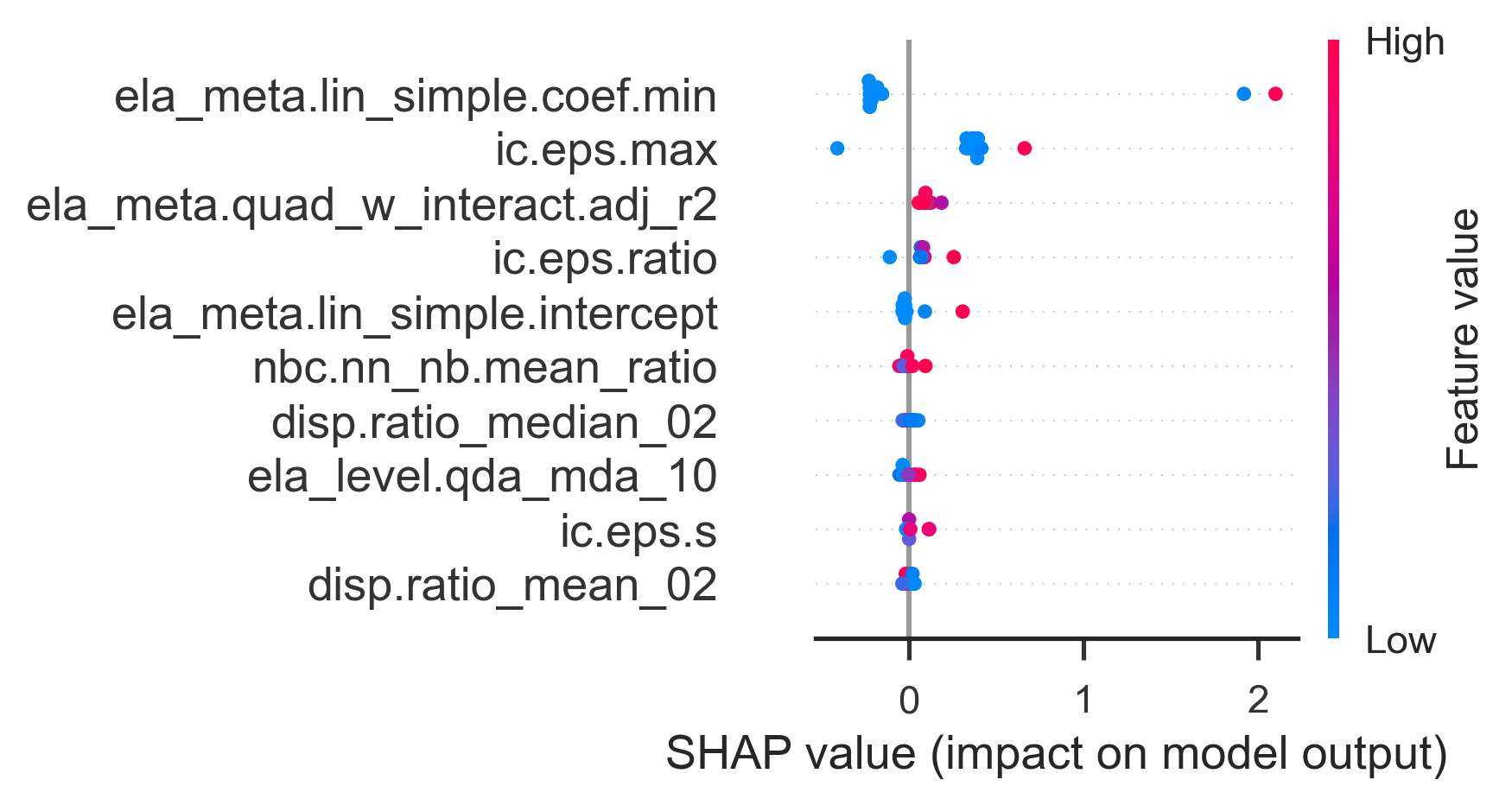}  
  \caption{Second fold (poor, good).}
  \label{fig:sub-fourth-shap}
\end{subfigure}
\caption{The 10 most important ELA features and their prediction influence for the test instances of the first and second fold for the (good, good) and (poor, good) clusters. Each point on the plot is a Shapley value for a feature and an instance. Its position on the y-axis is determined by the feature and on the x-axis by the Shapley value. The color represents the value of the feature from low to high.}
\label{fig:footprint-shap}
\end{figure*}

\noindent\textbf{Post-hoc explainable analyses.} Next, we provide post-hoc analysis to estimate the landscape features that make the problem instances easy to be solved $(good, good)$) and or challenging to be solved ($(poor, good)$). Figure~\ref{fig:footprint-shap} presents the 10 most important landscape features, for the first and second fold. The plots in this figure also depict both positive and negative relations with the target precision that is being predicted. The dots shown in the plots represent all instances from the selected folds. The ELA features are ordered based on their importance, with the most important feature being listed first. The color coding used reflects the magnitude of the ELA feature value, with higher values represented in red and lower values in blue. The effect of the ELA feature value on the target variable prediction can be seen by its horizontal placement. From the figure, it is obvious that the features for the $(good, good)$ cluster across both presented folds are overlapping for seven out of 10 ELA features and the same patterns of influence are presented. In the case of the $(poor, good)$ cluster, both folds are overlapping in eight out of 10 ELA features with similar patterns of influence. Comparing the $(good, good)$ vs. $(poor, good)$ in both folds separately, we can see that the overlapping is in a few ELA features, however, even the influence patterns of those which are overlapping are different (e.g., ic.eps.max, ela\_meta.lin\_simple.coef.min).

\begin{figure*}[htb]
\begin{subfigure}{.4\textwidth}
  \centering
  % include first image
  \includegraphics[width=.95\linewidth]{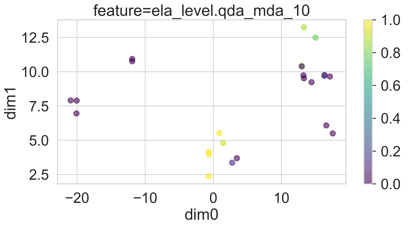}  
  \caption{First fold.}
  \label{fig:sub-first-shap-d}
\end{subfigure}
\begin{subfigure}{.4\textwidth}
  \centering
  % include second image
  \includegraphics[width=.95\linewidth]{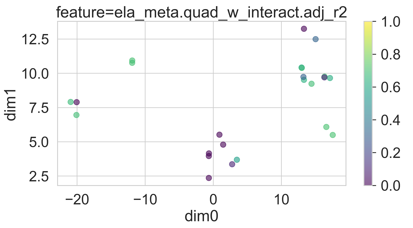}  
  \caption{First fold.}
  \label{fig:sub-second-shap-d}
\end{subfigure}

\leavevmode\newline
\vspace{-7mm}

\begin{subfigure}{.4\textwidth}
  \centering
  % include third image
  \includegraphics[width=.95\linewidth]{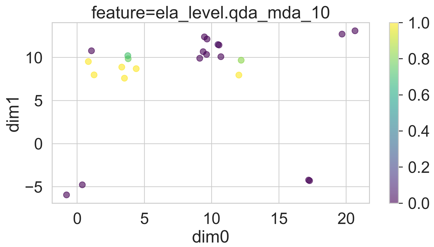}  
  \caption{Second fold.}
  \label{fig:sub-third-shap-d}
\end{subfigure}
\begin{subfigure}{.4\textwidth}
  \centering
  % include fourth image
  \includegraphics[width=.95\linewidth]{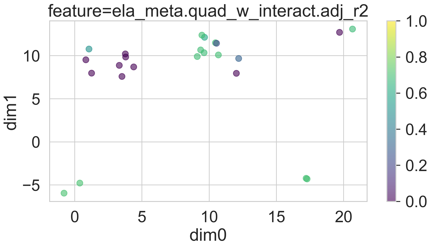}  
  \caption{Second fold.}
  \label{fig:sub-fourth-shap-d}
\end{subfigure}
\vspace{-2mm}
\caption{The distribution of two randomly selected (from the top 10) ELA features across the algorithm instance footprint. The color in the plots represents the normalized feature values.} 
\label{fig:footprint-shap_distirbution}

\end{figure*}

To go into more detail, we randomly selected two ELA features (\textit{ela\_level.qda\_mda\_10} and \textit{ela\_meta.quad\_w\_interact.adj\_r2}) and present their distributions across the algorithm instance footprint (see Figure~\ref{fig:footprint-shap_distirbution}). The distributions of the other ELA features are available in our GitHub repository [link omitted during the review]. 

In the case of the algorithm footprint generated on the first test fold, it is obvious that \textit{ela\_level.qda\_mda\_10} has higher values for the problem instances that belong to $(good, good)$ cluster and lower values for the problem instances that belong to $(poor, good)$ cluster. The opposite is true for the \textit{ela\_meta.quad\_w\_interact.adj\_r2}, where lower values are for problem instances that belong to $(good, good)$ and higher values are related to problem instances that belong to the $(poor, good)$ cluster.

For the algorithm footprint generated on the second test fold, the \textit{ela\_level.qda\_mda\_10} has higher values for the problem instances from the $(good, good)$ cluster and lower values for those that belong to the $(poor, good)$ cluster. In the case of the \textit{ela\_level.qda\_mda\_10} feature, lower values are associated with the problem instances from the $(good, good)$ and medium values are related to the instances from the $(poor, good)$ cluster.

\begin{figure*}
\begin{subfigure}{.4\textwidth}
  \centering
  % include first image
  \includegraphics[width=.95\linewidth]{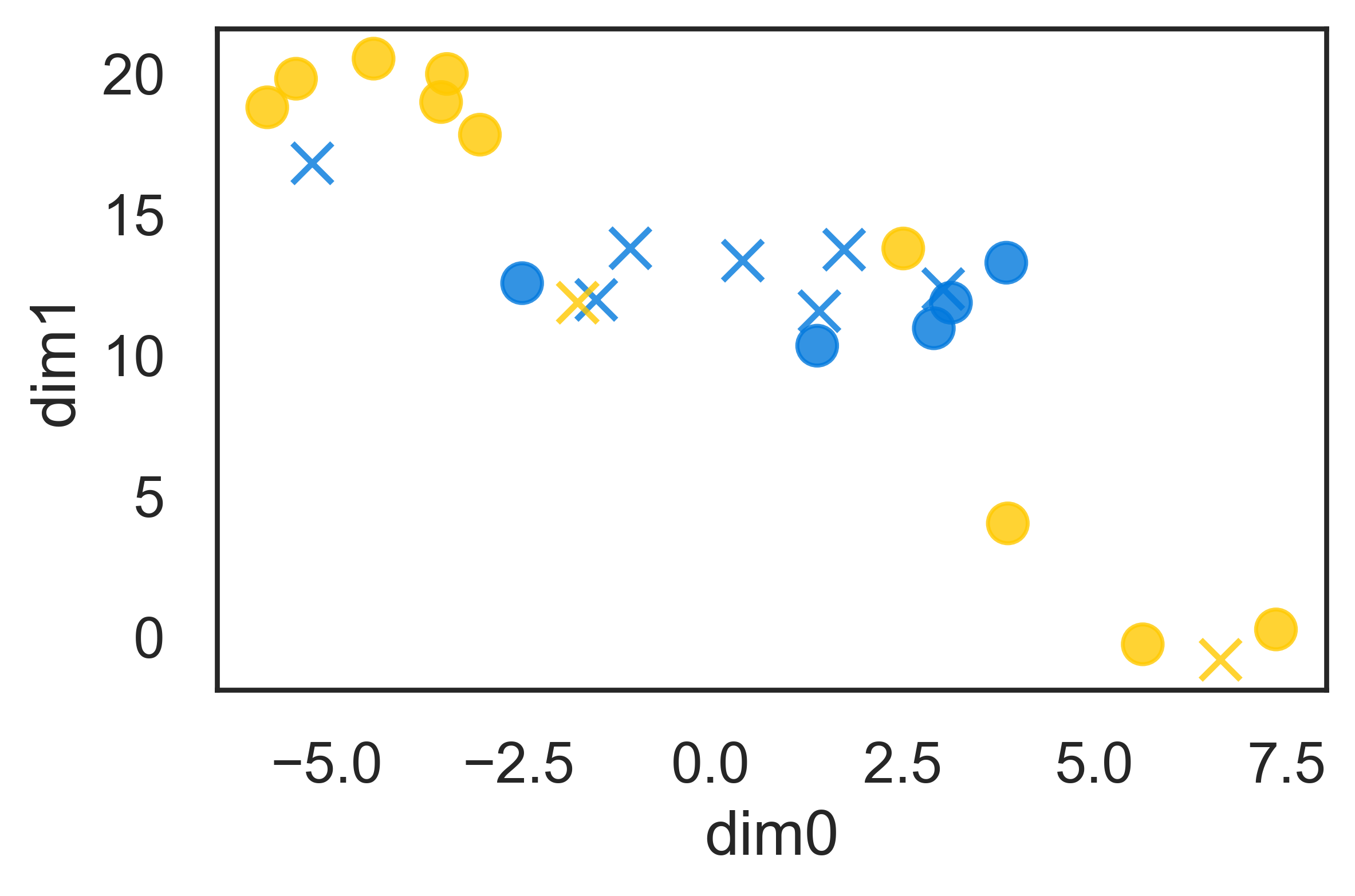}  
  \caption{First fold of KNN.}
  \label{fig:sub-fifth-shap-knn}
\end{subfigure}
\begin{subfigure}{.4\textwidth}
  \centering
  % include second image
  \includegraphics[width=.95\linewidth]{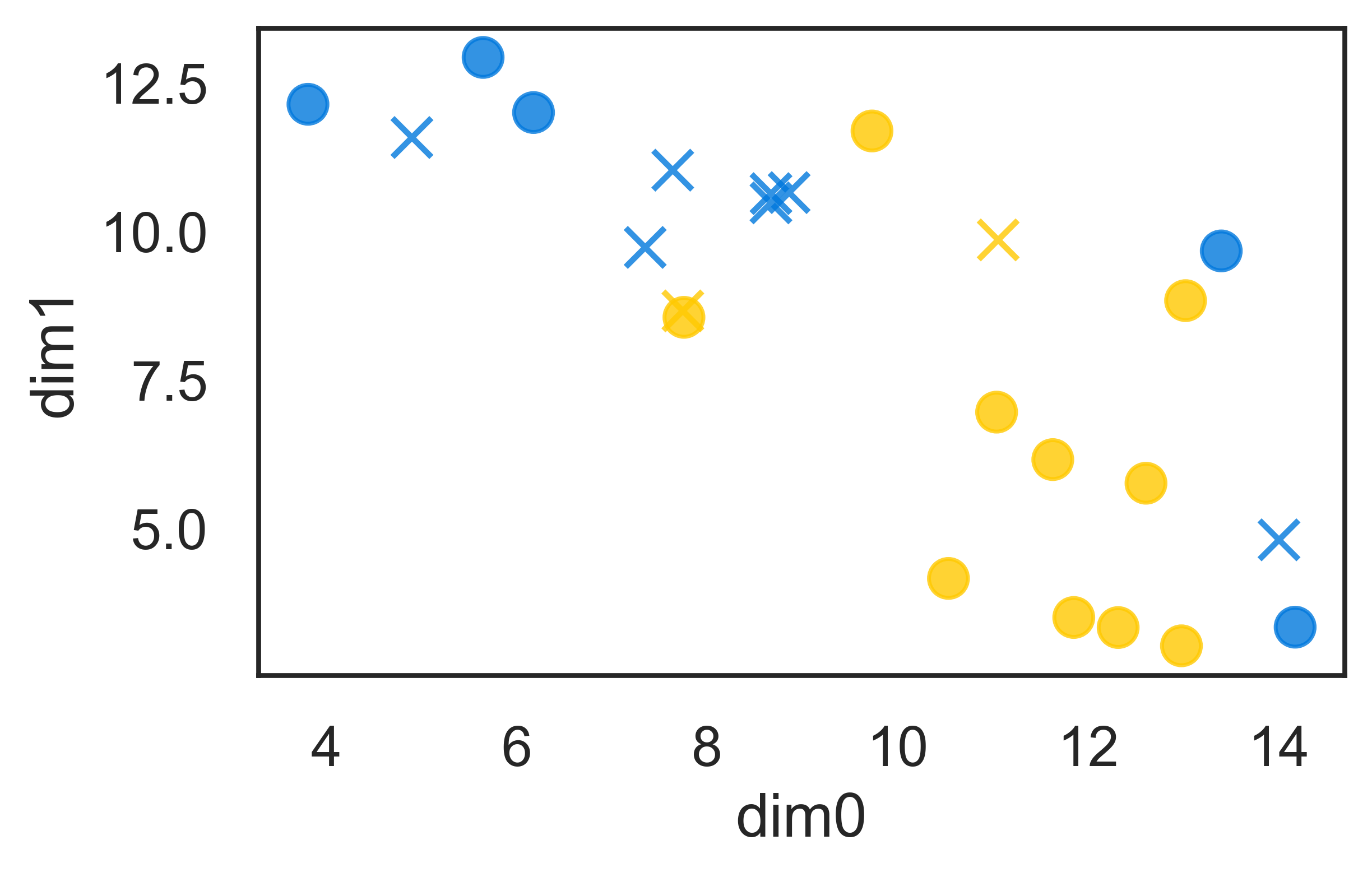}  
  \caption{First fold of SVM.}
  \label{fig:sub-sixth-shap-svm}
\end{subfigure}
\vspace{-2mm}
\caption{2D UMAP visualization of the algorithm footprints obtained with the deterministic clustering on the test portion of the first fold. The tolerance error for the KNN and SVM model is within 15\%.
The \textit{\textcolor{blue}{blue}} color represents regions of good algorithm performance, and the \textit{\textcolor{yellow}{yellow}} to regions of poor algorithm performance. 
The marker shape corresponds to good (O) and poor (X) ML model performance.}

\label{fig:footprint-KNN-SVM}
\end{figure*}

\noindent\textbf{DE1 footprints generated by different ML models.} Figure~\ref{fig:footprint-KNN-SVM} presents the footprints generated for DE1 and the first fold by using different ML predictive models, KNN and SVM. The tolerance error for both ML models is set at 15\% in order to benchmark the footprints with the footprint generated by the RF model. Comparing the footprints with the footprint presented in Figure~\ref{fig:sub-first}, it is obvious that the distribution of the problem instances in the space is different since the meta-representations are model-specific and generated by using different supervised ML models. Table~\ref{t:distribution} presents the distribution of the BBOB problem instances across the footprints generated for all folds by RF, KNN, and SVM. All ML models provided similar results on the first fold which is also visible by the distribution of the problem instances across all four deterministic clusters. No matter which model is used, the clusters are almost the same. The distributions of the problem instances slightly change for the other remaining folds, which indicates the model-specific aspect and the importance of selecting a good-performing ML model. However, similar distributions that are achieved across the folds support the fact that the proposed methodology can be used to analyze the strengths and weaknesses of algorithm instance behavior. Currently, the explanations (i.e., important landscape features) are model-specific, they depend on the selection of the supervised regression model. In the future, we are planning to find the intersection of important landscape features across footprints generated with different ML models in order to go through a model-agnostic approach. 

\begin{table*}[!htb]
\caption{Distribution of the BBOB problem instances across the deterministic clusters for each fold.}
\label{t:distribution}
\centering
\vspace{-2mm}
\resizebox{.8\textwidth}{!}{%
\begin{tabular}{llllll}
\hline
model & fold number &     (good, good) &  (good, poor) & (poor, good) &   (poor, poor) \\
\hline
RF & 1 & 16, 19, 20, 21, 22 & 1, 2, 5, 14, 17, 18, 23 & 3, 4, 6, 7, 8, 9, 10, 11, 12, 15, 24 & 13 \\
KNN & 1           &      2, 16, 18, 19, 20, 21, 23 &  1, 5, 14, 17, 22 &     4, 6, 7, 8, 9, 10, 11, 12, 15, 24 &            3, 13 \\
SVM & 1           &  16, 19, 20, 21, 22 &              1, 2, 5, 14, 17, 18, 23 &  3, 4, 6, 7, 8, 9, 10, 11, 12, 24 &                     13, 15 \\
\hline
RF & 2 & 19, 20, 21 & 1, 2, 5, 14, 17, 22, 23 & 3, 4, 6, 7, 8, 9, 10, 11, 12, 13, 15, 16, 24 & 18 \\
KNN & 2           &              5, 17, 19, 20, 23 &  1, 2, 14, 21, 22 &  3, 4, 6, 7, 8, 9, 10, 11, 12, 16, 24 &       13, 15, 18 \\
SVM & 2           & 19, 21 & 1, 2, 5, 14, 17, 20, 22, 23 & 6, 8, 9, 12, 15, 16, 18, 24 & 3, 4, 7, 10, 11, 13 \\
\hline
RF & 3 & 19, 20, 21, 22 & 1, 2, 5, 14, 16, 17, 18, 23 & 3, 4, 6, 8, 9, 12, 13, 15, 24 & 7, 10, 11 \\
KNN & 3           &  1, 16, 18, 19, 20, 21, 22, 23 &      2, 5, 14, 17 &               6, 8, 9, 10, 12, 13, 24 &  3, 4, 7, 11, 15 \\
SVM & 3           &              6, 22 &  1, 2, 5, 14, 17, 18, 19, 20, 21, 23 &                     9, 12, 13, 24 &  3, 4, 6, 7, 8, 10, 11, 15 \\
\hline
RF & 4 & 5, 16, 18, 19, 20, 21, 22 & 1, 2, 7, 14, 17, 23 & 3, 4, 6, 8, 9, 10, 12, 13, 15, 24 & 11 \\
KNN & 4           &   1, 7, 16, 19, 20, 21, 22, 23 &  2, 5, 14, 17, 18 &                4, 6, 8, 9, 10, 12, 24 &    3, 11, 13, 15 \\
SVM & 4           &          16, 20, 21 &   1, 2, 5, 7, 14, 17, 18, 19, 22, 23 &                  6, 9, 11, 13, 24 &        3, 4, 8, 10, 12, 15 \\
\hline
RF & 5 & 19, 20, 21 & 1, 2, 5, 7, 14, 16, 17, 22, 23 & 6, 8, 9, 11, 12, 13, 15, 24 & 3, 4, 10, 18 \\
KNN & 5           &  1, 14, 16, 19, 20, 21, 22, 23 &       2, 5, 7, 17 &            4, 6, 8, 9, 11, 12, 15, 24 &    3, 10, 13, 18 \\
SVM & 5           & 5, 16, 19, 21, 22 &            1, 2, 7, 14, 17, 20, 23     & 3, 8, 9, 10, 11, 12, 18, 24 & 4, 6, 13, 15 \\
\hline
\end{tabular}%
}
\end{table*}

\textbf{Discussion.} 
Our methodology is proposed as an exploratory tool for analyzing and understanding the strengths and weaknesses of a selected algorithm instance. It cannot be used to benchmark different algorithm instances based on their footprints. This comes with the fact that the footprint of each algorithm is generated using explanations of a single-target regression model, which means that each footprint is generated in its own vector space. As future work, we are going to purpose footprints that can be used for transparent benchmarking where the algorithm performance prediction will be done as multi-task learning where the performance of different algorithm instances will be treated as multiple learning tasks that are solved at the same time while exploiting commonalities and differences across tasks. 

In our experiments, the predefined target precision for the algorithm instance performance, $t$, and the tolerance percentage of the ML error, $p$, were set only for illustration purposes. Those parameters can be set by the researchers/users depending on their application requirements. They can also be changed in order to explore the sensitivity of the footprints. In addition, when training the ML predictive model for algorithm instance performance prediction, more advanced techniques based on AutoML~\cite{Feurer2019} are recommended to find the best ML predictive model (i.e., in our experiments we use RF with default parameters for illustration purposes).

We illustrate the generation of the algorithm instance footprint using 10$D$ problem instances. In the future, footprints for the same algorithm instance can be generated for a different dimension to explore the distribution of the problem instances across the footprint and also to explore the importance of the landscape features when the dimension of the problem instances increases.

We use ELA features that provide some information but are still low-interpretable. In the future, this methodology can also be performed using another portfolio of landscape features (e.g., high-level features) in order to provide more human-interpretable explanations. Currently, the explanations are model-specific, they depend on the selection of the supervised regression model. In the future, we are planning to generate footprints for the same algorithm instance with regard to different regression models and try to find the intersection between them in order to go through a model-agnostic approach. 

While sharing philosophical foundations, the proposed methodology differs in approach to ISA~\cite{munoz2017performance}. For example, ISA attempts to give both an indication of the diversity of the benchmark instances and serve as a tool for developing hypotheses on the strengths and weaknesses of the algorithms. For this, ISA finds a common space for all algorithms, by selecting features that, after a linear projection, are the most predictive of performance on average across the portfolio. 
Moreover, ISA constructs and measures the footprints as regions of the space through a combination of clustering and geometric methods~\cite{smith-miles2022instance}, to account for the diversity of the instances. 
Nevertheless, no investigation of multiple algorithm instances through ISA has been made in SOO. 
The interpretation in ISA is human-driven, with an analysis of the visualizations by the researchers being an important step, with a standardized post-hoc analysis not being part of the ISA yet. A comparison between the results of the proposed methodology and ISA is left for further research.

\section{Conclusions}
\label{sec:conclusion}

In the context of black-box optimization, it's crucial to comprehend the reasons why an algorithm instance works well on some problem instances and fails on others. We introduce a method for creating an algorithm instance footprint, which consists of dividing the problem instances into two sets: those that are easily solvable and those that are challenging. This is achieved by linking the algorithm's behavior to the properties of the problem landscape, providing explanations of why some problem instances are easier or harder. Our methodology employs meta-representations, which embed the properties of the problem instances and the performance of the algorithm into the same vector space. These meta-representations are obtained through training a supervised machine learning regression model and examining feature importance. Additionally, deterministic clustering of the meta-representations reveals regions of good and poor algorithm performance, along with an explanation of which landscape properties are responsible. This analysis enables us to gain insight into the strengths and weaknesses of the algorithm instance and move away from treating it as a black-box.

\vspace{1ex}
{\footnotesize{
% \begin{acks}
\textbf{Acknowledgments: }
The authors acknowledge the support of the Slovenian Research Agency through program grant No. P2-0098 and P2-0103  project grants N2-0239 and J2-4460, and a bilateral project between Slovenia and France grant No. BI-FR/23-24-PROTEUS-001 (PR-12040), as well as the EC through grant No. 952215 (TAILOR). Our work is also supported by ANR-22-ERCS-0003-01 project VARIATION.
% \end{acks}
}}

% \bibliographystyle{ACM-Reference-Format}

\end{document}